\documentclass[10pt,twocolumn,letterpaper]{article}

\usepackage{iccv}              
\usepackage[accsupp]{axessibility}
\usepackage{multirow}
\usepackage[table,xcdraw]{xcolor}
\usepackage{pifont}
\usepackage{tablefootnote}
\usepackage[x11names]{xcolor}
\usepackage{makecell}

\renewcommand{\ie}{\emph{i.e.,}}

\newcommand{\figref}[1]{Fig.~\ref{#1}}
\newcommand{\tabref}[1]{Tab.~\ref{#1}}

\definecolor{mygray}{gray}{.9}

\newcommand{\name}{PIG}

\definecolor{iccvblue}{rgb}{0.21,0.49,0.74}
\usepackage[pagebackref,breaklinks,colorlinks,allcolors=iccvblue]{hyperref}

\def\paperID{12350} 
\def\confName{ICCV}
\def\confYear{2025}

\title{Hybrid-Tower: Fine-grained Pseudo-query Interaction and Generation for Text-to-Video Retrieval}

\author{Bangxiang Lan$^{1}$ \thanks{Work performed as an intern at Tencent. (bangxiang@ruc.edu.cn)} \quad Ruobing Xie$^{2}$ \quad Ruixiang Zhao$^{1}$ \\ Xingwu Sun$^{2}$ \quad Zhanhui Kang$^{2}$ \quad Gang Yang$^{1}$\thanks{Corresponding author: Gang Yang (yanggang@ruc.edu.cn)} \quad Xirong Li$^{1}$ \\
$^{1}$Renmin University of China \quad $^{2}$Large Language Model Department, Tencent
\\
{\small \tt \url{https://lbx73737373.github.io/PIG-ProjectPage/}}
}

\begin{document}
\maketitle
\begin{abstract}
The Text-to-Video Retrieval (T2VR) task aims to retrieve unlabeled videos by textual queries with the same semantic meanings. Recent CLIP-based approaches have explored two frameworks: Two-Tower versus Single-Tower framework, yet the former suffers from low effectiveness, while the latter suffers from low efficiency. In this study, we explore a new Hybrid-Tower framework that can hybridize the advantages of the Two-Tower and Single-Tower framework, achieving high effectiveness and efficiency simultaneously. 
We propose a novel hybrid method, Fine-grained Pseudo-query Interaction and Generation for T2VR, \ie \name{}, which includes a new pseudo-query generator designed to generate a pseudo-query for each video. This enables the video feature and the textual features of pseudo-query to interact in a fine-grained manner, similar to the Single-Tower approaches to hold high effectiveness, even before the real textual query is received. 
Simultaneously, our method introduces no additional storage or computational overhead compared to the Two-Tower framework during the inference stage, thus maintaining high efficiency.
Extensive experiments on five commonly used text-video retrieval benchmarks demonstrate that our method achieves a significant improvement over the baseline, with an increase of 
$1.6\% \sim 3.9\%$ in R@1. Furthermore, our method matches the efficiency of Two-Tower models while achieving near state-of-the-art performance, highlighting the advantages of the Hybrid-Tower framework.
\end{abstract}
\section{Introduction}
\label{sec:intro}

\begin{figure}[tbp!]
    \centering
    \begin{subfigure}[b]{0.45\linewidth}
        \centering
        \includegraphics[height=4.5cm,keepaspectratio]
        {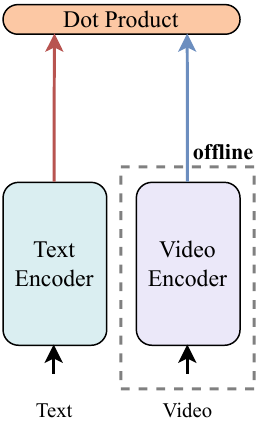}
        \caption{
        \centering Two-Tower: Low effectiveness, High efficiency.
        }
        \label{fig:sub-two-tower}
    \end{subfigure}
    \hfill
    \begin{subfigure}[b]{0.45\linewidth}
        \centering
        \includegraphics[height=4.5cm,keepaspectratio]{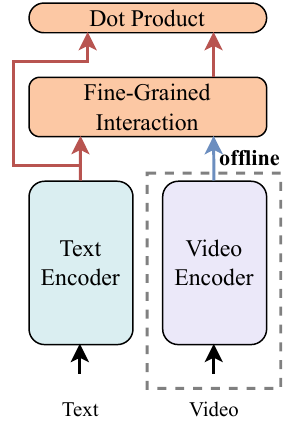}
        \caption{\centering Single-Tower: High effectiveness, Low efficiency. 
        }
        \label{fig:sub-single-tower}
    \end{subfigure}

    \vspace{1em}

    \begin{subfigure}[b]{0.75\linewidth}
        \centering
        \includegraphics[height=4.5cm,keepaspectratio]{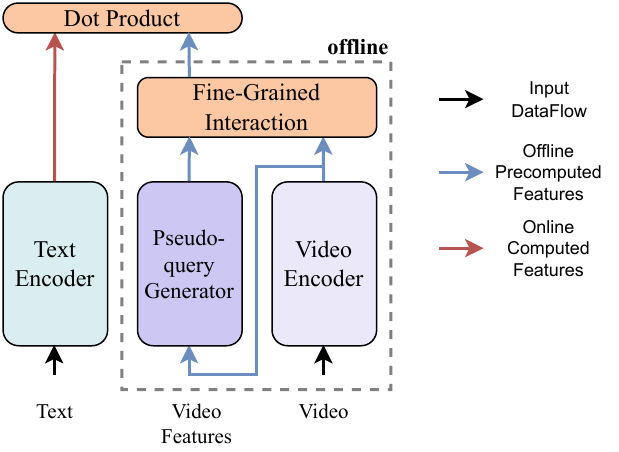}
        \caption{
            Hybrid-Tower: High effectiveness, High efficiency
        }
        \label{fig:sub-hybrid-tower}
    \end{subfigure}

    \caption{Illustration of two conventional text-to-video retrieval model frameworks: (a) Two-Tower, (b) Single-Tower and (c) our proposed Hybrid-Tower. The modules enclosed in dashed boxes indicate that they can be precomputed offline, so the bottleneck during online inference lies in other modules.}
    \label{fig:frameworks}
    \vspace{-13pt}
\end{figure}

Recently, Text-to-Video Retrieval (T2VR) has seen significant improvements due to the advancement of the Contrastive Language-Image Pre-Training (CLIP) \cite{icml21clip} model. Researchers have been putting effort into fully exploiting the potential of CLIP in the video-text domain. Given the temporal nature of video, some studies incorporate a time modeling module into CLIP \cite{Neurocomputing2022clip4clip, eccv2022ts2, cvpr2023stan}, some focus on improving performance by considering fine-grained frame-text similarities \cite{mm22xclip, cvpr2022xpool, iccv2023promptswitch, cvpr2023banzhaf, iccv2023diffusionret}, and others attempt to introduce external knowledge to enhance CLIP \cite{cvpr2023cap4video, cvpr2024teachclip, eccv2024kdpror}. Nevertheless, these works can be broadly categorized into two classic frameworks: Two-Tower and Single-Tower, as shown in \figref{fig:sub-two-tower} and \figref{fig:sub-single-tower}, respectively. 
Both frameworks consist of a video encoder and a text encoder. After extracting text and video features, their similarity is computed using a simple dot product. Single-Tower methods introduce an additional fine-grained interaction module, such as similarity fusion in X-CLIP \cite{mm22xclip} or feature fusion in XPool \cite{cvpr2022xpool}. This fine-grained interaction enables Single-Tower models to derive more accurate video representations, leading to improved retrieval performance compared to Two-Tower models on average.

However, in real-world retrieval scenarios, the interaction module in Single-Tower models become the bottleneck for retrieval speed. This is because, in Single-Tower models, video feature extraction is coupled with text queries; that is, for a given text query, every video in the dataset must undergo feature fusion in an additional interaction module. In contrast, Two-Tower models rely solely on a simple dot product, allowing the final video representations to be computed offline and stored in advance, making retrieval significantly faster than in Single-Tower models. The computational offline components in each framework are illustrated in \figref{fig:frameworks}. In summary, while Single-Tower models achieve higher retrieval accuracy through fine-grained interaction, they suffer from lower efficiency.

Given the limitations of existing frameworks, we ask: \textit{Is there a new framework that can simultaneously achieve high retrieval speed and high retrieval accuracy?} In this paper, we propose a new Hybrid-Tower framework, illustrated in \figref{fig:sub-hybrid-tower}, which contains a novel pseudo-query generator. This generator enables acquisition of pseudo-queries related to different videos before the arrival of real text queries, allowing the video features to perform a fine-grained interaction with pseudo-text queries, similar to Single-Tower models. We propose a method under Hybrid-Tower framework: Fine-grained \textbf{\underline{P}}seudo-query \textbf{\underline{I}}nteraction and \textbf{\underline{G}}eneration for T2VR, namely \textbf{\name{}}. It tackles this problem through two core components: 
(i) \emph{pseudo-query generator} that leverages multi-grained visual features and an informative token selection mechanism to generate discriminative pseudo-queries. 
(ii) \emph{pseudo-interaction fusioner} that performs fine-grained pseudo-query interaction to enhance the final video representation. Experimental results on five standard datasets show that \name{} could achieve near-SOTA performance while maintaining significantly higher efficiency. 
Furthermore, the new proposed framework could encourage further investigation in the Text-to-Video retrieval community.
In summary, our contributions are as follows:
\begin{itemize}
    \item We introduce a novel framework for T2VR: Hybrid-Tower, which combines and obtains the effectiveness of Single-Tower models and the efficiency of Two-Tower models.
    \item We propose the \name{} method under the Hybrid-Tower framework, which generates pseudo-queries to enable fine-grained feature fusion with videos to realize effective interaction. The core of \name{} is a causal attention-powered pseudo-query generator with an informative token selection (ITS) module and a fine-grained pseudo-interaction fusioner. 
    \item Extensive experiments conducted on five video retrieval datasets, \ie  MSRVTT-1ka \cite{eccv2018joint}, MSRVTT-3k \cite{cvpr2016msrvtt}, MSVD \cite{acl2011msvd}, VATEX \cite{iccv2019vatex}, and DiDeMo \cite{emnlp2018didemo}, demonstrate the effectiveness and efficiency of our method. Our proposed method achieves near-SOTA performance of Single-Tower and the SOTA efficiency of Two-Tower models, simultaneously. 
\end{itemize}


\section{Related work}
\label{sec:related_work}

\subsection{Effective T2VR}
The majority of existing literature focuses on obtaining more effective video and text representations. Based on whether text and video encoders are jointly trained with the similarity calculation module, current approaches can be categorized into two groups: feature re-learning and CLIP-based end-to-end methods within a Single-Tower architecture, as illustrated in \figref{fig:sub-two-tower}.

Feature re-learning methods typically use pre-trained 2D-CNNs \cite{pami2021de, acmmm2019w2vv++}, 3D-CNNs \cite{ce, icmr2018learning}, or combinations thereof \cite{eccv2020mmt, eccv2022laff} to generate initial video features. Text encoding usually employs non-trainable bag-of-words models \cite{acmmm2019w2vv++} or pre-trained encoders like Word2Vec \cite{tmm2018predicting}, BERT \cite{acmmm2019w2vv++}, and GPT \cite{cvpr2021teachtext}. Both video and text features are then projected into a shared latent space to evaluate their relevance based on their distances. Although improvements such as novel feature fusion or enhancement modules have been proposed \cite{cvpr2021teachtext, eccv2020mmt, eccv2022laff}, the performance of feature re-learning methods remains largely dependent on initial feature quality.

With advances in image-language models, CLIP-based methods have emerged as more effective for Text-to-Video Retrieval (T2VR), outperforming traditional feature re-learning methods. A critical component of these methods is the fine-grained text-video interaction within the Single-Tower framework, enhancing video representations. Interaction modules have progressively evolved from simple video-sentence alignment \cite{Neurocomputing2022clip4clip} to more detailed frame-sentence \cite{arxiv2022drl, cvpr2022xpool, iccv2023uatvr}, frame-word \cite{mm22xclip, cvpr2023banzhaf}, and patch-word alignments \cite{cvpr2023prost, cvpr2023pidro, iccv2023ucofia}. However, increasing complexity in these interaction modules significantly reduces efficiency. As shown in \tabref{tab:alldatasets}, recent Single-Tower models such as UCoFia incur tens-of-thousands-fold efficiency costs for marginal performance improvements, presenting a major bottleneck for T2VR advancement.

\subsection{Efficient T2VR}

Depending on how the term “efficiency” is interpreted, existing studies can be categorized into three groups: (i) training efficiency, aiming to train T2VR models in a parameter-efficient fine-tuning (PEFT) manner \cite{aaai2024dgl, cvpr2023vop, cvpr2024mvadapter}; (ii) feature extraction efficiency, focusing on reducing the size of the video encoder \cite{cvpr2023clipping, sigir2022centerclip}; and (iii) serving efficiency, which aims to boost speed in a real retrieval scenario, where video features are pre-cached. Our study falls into the third category.

To enhance serving speed, efficient CLIP-based methods typically retain only the simple dot product operation, fitting into the Two-Tower framework as illustrated in \figref{fig:sub-single-tower}. PromptSwitch \cite{iccv2023promptswitch} is among the first efficient CLIP-based T2VR methods, introducing a prompt cube into CLIP to iteratively model pair-to-pair temporal frame interactions. EERCF \cite{AAAI2024EERCF} adopts a two-stage retrieval strategy, first utilizing coarse-grained video representations to recall the top-k candidates rapidly, which are then reranked using finer-grained representations. TeachCLIP employs external fine-grained teachers, leveraging knowledge distillation, injecting the frame-sentence dependencies from the teacher into an Attentional frame-Feature Aggregation (AFA) module.

In contrast to these methods, \name{} adopts a hybrid perspective. Specifically, it belongs to our novel Hybrid-Tower framework (see \figref{fig:sub-hybrid-tower}), which retains the Two-Tower structure during serving while enhancing effectiveness through pseudo-query generation and pseudo-interactions with pre-cached video representations.

\section{Method}
\label{sec:method}

We propose a novel \name{} framework that involves pre-identifying potential text queries associated with target videos to achieve an optimal balance between computational efficiency and retrieval effectiveness.

\begin{figure*}[!htbp]
    \centering
    \includegraphics[width=\textwidth]{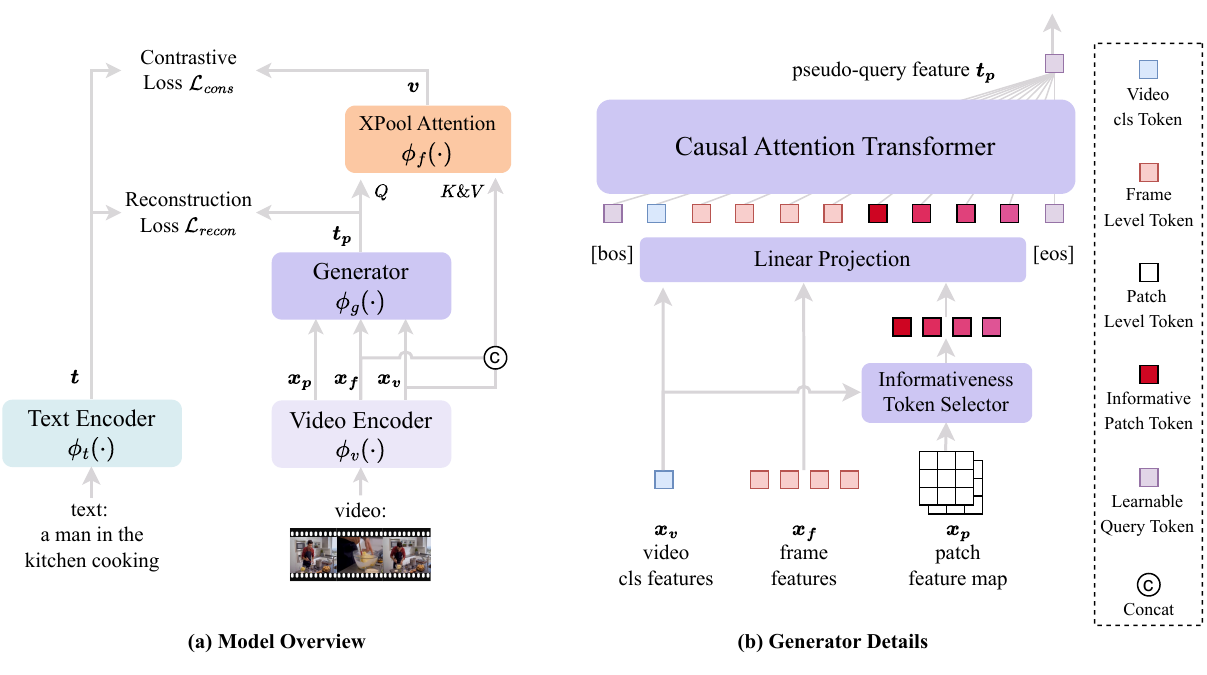}
    \caption{An overview of our proposed \name{} for text-to-video retrieval: (a) the overall structure of our model. (b) A detailed illustration of our proposed pseudo-query generator.}
    \label{fig:overview}
\end{figure*}

\subsection{Preliminaries}
\label{subsec:preliminaries}
Given a multi-modality dataset consisting of $N_v$ video clips and $N_t$ text, \ie $\mathcal{D} = \{v_i, t_j\}$, where each video clip has one or more corresponding text queries. The goal of text-to-video retrieval (T2VR) is to rank the videos in the video gallery based on their relevance to a given text query. The method for T2VR typically employs a text encoder $\phi_t(\cdot)$ and a video encoder $\phi_v(\cdot)$ to abstract the video and the text into a shared cross-modal feature space: 
\begin{equation}
    \boldsymbol{v} = \phi_v(v); \boldsymbol{t} = \phi_t(t),
\end{equation}
where $\boldsymbol{v}, \boldsymbol{t} \in \mathbb{R}^d$ and $d$ denotes the embedding dimension of the feature space. The cosine similarity $s(\boldsymbol{t}, \boldsymbol{v}) = \frac{\boldsymbol{t}\cdot \boldsymbol{v}}{\| \boldsymbol{t} \| ~ \|\boldsymbol{v}\|}$ is adopted to measure the distance between a video clip and a text query. In training, given a batch of $B$ text-video pairs, a widely-adopted symmetric contrastive loss \ie InfoNCE \cite{oord2018representation, sohn2016improved, icml21clip, mm22xclip} is used to optimize the model:
\begin{equation}
    \mathcal{L}_{t2v} = -\frac{1}{B}\sum_{i}^{B} \mathrm{log} \frac{\mathrm{exp}(s(\boldsymbol{t}_i, \boldsymbol{v}_i) \cdot \tau)}{\sum_{j=1}^{B} \mathrm{exp} (s(\boldsymbol{t}_i, \boldsymbol{v}_j) \cdot \tau)},
    \label{eq:t2v}
\end{equation}
\begin{equation}
    \mathcal{L}_{v2t} = -\frac{1}{B}\sum_{i}^{B} \mathrm{log} \frac{\mathrm{exp}(s(\boldsymbol{v}_i, \boldsymbol{t}_i) \cdot \tau)}{\sum_{j=1}^{B} \mathrm{exp} (s(\boldsymbol{v}_i, \boldsymbol{t}_j) \cdot \tau)}, 
    \label{eq:v2t}
\end{equation}
\begin{equation}
    \mathcal{L}_{cons} = \frac{1}{2} (\mathcal{L}_{t2v} + \mathcal{L}_{v2t}),
    \label{eq:t2v-v2t}
\end{equation}
where $\tau$ is a learnable temperature scaling factor.

One of the recent trends in CLIP-based T2VR is to strengthen temporal modeling ability for video encoder of CLIP \cite{Neurocomputing2022clip4clip, eccv2022ts2, cvpr2023stan, cvpr2023prost, cvpr2024mvadapter, iclr2023clipvip}. Among these models, CLIP-ViP \cite{iclr2023clipvip} stands out as a powerful approach that introduces video proxy tokens and ViP-guided attention mechanism. Thus, we choose CLIP-ViP as the backbone encoder of our approach.

\subsection{Overall Framework}
\label{subsec:framework}
In this work, we propose a novel Hybrid-Tower framework for T2VR and introduce a new model, \name{}, which features fine-grained pseudo-query interaction and generation, as illustrated in \figref{fig:overview}. \name{} can generate pseudo-query features from visual inputs to perform a fine-grained pseudo interaction of query-video in a Single-Tower manner ahead of the arrival of real textual queries. After the pseudo interaction, the learned video representations are retrieved by queries in an efficient Two-Tower manner. Since we can ``pre-fuse'' the pseudo query information into video representations, the effectiveness of Single-Tower methods and the efficiency of Two-Tower methods can be simultaneously assured.
There are four essential components in \name{}: a text encoder $\phi_t(\cdot)$, a video encoder $\phi_v(\cdot)$, a pseudo-query generator $\phi_g(\cdot)$ and a pseudo interaction fusioner $\phi_f(\cdot)$ as follows.


\subsection{Text and Video Encoders}
Given a video clip $v$ comprising $m$ frames $\{f_1, \dots, f_m\}$, the video encoder $\phi_v(\cdot)$ feeds the whole frame sequence into the visual encoder of CLIP \cite{icml21clip}, \ie a Vision Transformer(ViT) to produce visual features. 
The CLIP-ViP is applied as the backbone of our method with just a minor modification: we add the frame-level \texttt{[cls]} token to the ViT of CLIP-ViP, enabling it to provide extra frame-level visual outputs.
With the help of video proxies in CLIP-ViP \cite{iclr2023clipvip}, we can obtain multi-grained visual features by selecting the video-level \texttt{[cls]}, frame-level \texttt{[cls]} and patch-level tokens from the output of the last ViT layer, denoted as $\boldsymbol{x_v^{’}}$, $\boldsymbol{x_f^{’}}$, and $\boldsymbol{x_p^{’}}$, respectively. These features are then projected into the cross-modal space through a visual linear projection layer in CLIP. These multi-grained features are extracted for the next pseudo-query generation. More formally, the procedure can be expressed as follows:
\begin{equation}
\left\{ \begin{array}{ll}
 \{f_1, \ldots, f_m\} & \leftarrow \mbox{video-to-frames}(v),\\
 \boldsymbol{x_v^{'}}, \boldsymbol{x_f^{'}}, \boldsymbol{x_p^{'}} & \leftarrow \mbox{ViT}(\{f_1, \ldots, f_m\}),\\
 \boldsymbol{x_v}, \boldsymbol{x_f}, \boldsymbol{x_p} & \leftarrow \mbox{Linear}(\boldsymbol{x_v^{'}}, \boldsymbol{x_f^{'}}, \boldsymbol{x_p^{'}}),
       \end{array} \right.
\end{equation}
where $\boldsymbol{x_v} \in \mathbb{R}^{4 \times d}$, $\boldsymbol{x_f} \in \mathbb{R}^{m \times d}$, and $\boldsymbol{x_p} \in \mathbb{R}^{m \times n \times d}$, with $m$ representing the number of frames, $n$ the sequence length of patch tokens and $4$ the number of video-level tokens.

The text encoder is the standard CLIP text encoder, which extracts text features $\boldsymbol{t} = \phi_t(t)$, where $\boldsymbol{t} \in \mathbb{R}^d$.

\subsection{Pseudo-Query Generator}
\label{subsubsec:generator}

After video encoding, we design a pseudo-query generator that takes multi-grained visual features as input to generate(or reconstruct) fine-grained and discriminative pseudo text query $\boldsymbol{t_p}$, as $\boldsymbol{t_p} = \phi_g( [\boldsymbol{x_v}, \boldsymbol{x_f}, \boldsymbol{x_p}])$, where $[\cdot, \cdot]$ denotes concatenation, $\boldsymbol{t_p} \in \mathbb{R}^d$ and $[\boldsymbol{x_v}, \boldsymbol{x_f}, \boldsymbol{x_p}] \in \mathbb{R}^{(4+m+m\times n) \times d}$. 

Pseudo-query generator is crucial for generating discriminative pseudo queries of text that can enhance video representations via multi-modal feature fusion. 
We break this down into two parts: selecting the input features of the generator, and designing the architecture of the generator.

\begin{figure}[tbp!]
    \centering
    \includegraphics[width=\linewidth]{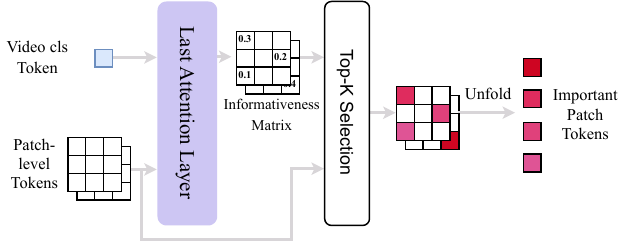}
    \caption{\textbf{Informativeness Token Selector.} We calculate the attention distribution of video cls token over all patch-level tokens to obtain an informativeness matrix. Referring to the informativeness matrix, we can select the most important top-k patch tokens.}
    \label{fig:topk}
    \vspace{-3pt}
\end{figure}

\textbf{Informativeness Token Selection.} 
Generating or reconstructing a discriminative pseudo-query from video information is highly challenging, particularly in our cross-modal generation scenario. To maximize the use of multi-grained video features \cite{iccv2023uatvr, iccv2023ucofia}, we feed video-level, frame-level, and patch-level features into our generator. Due to the high redundancy and noise of patch-level features, they should be concentrated. 
To this end, we propose a simple yet effective Informativeness Token Selector(ITS) module, as shown in \figref{fig:topk}. Due to the attention sparsity in ViT, where only a few tokens provide meaningful information (with background tokens even hindering performance)\cite{iccv2023umt}, it is natural to utilize this mechanism to highlight the important patch tokens. Specifically, given the first video-level \texttt{[cls]} token $\boldsymbol{x_v} \in \mathbb{R}^{1\times d}$ and all the patch-level tokens $\boldsymbol{x_p} $
, we calculate the attention scores of the video encoder's last attention layer:
\begin{equation}
\left\{ \begin{array}{ll}
Q_v^h & \leftarrow \boldsymbol{x_v}W_Q^h, \quad
K_p^h  \leftarrow \boldsymbol{x_p}W_K^h, \\
S_h & \leftarrow \mbox{softmax}(\frac{Q_v^h (K_p^h)^{\top}}{\sqrt{d/n}}), \\
S & \leftarrow \mathop{\max}\limits_{h} S_h,
       \end{array} \right.
\end{equation}
where $W_Q^h$ and $W_K^h$ are the linear projections in the $h$-th head. The informativeness matrix $S \in [0, 1]^{m \times n}$ represents the importance level of each video patch token,  obtained by applying a max operation across the multi-heads. Referring to the index of top-k values in the informativeness matrix, we select the top-k most informative patch tokens. After unfolding, we obtain the important patch-level tokens $\boldsymbol{x_{ip}} \in \mathbb{R}^{k \times d} $.  Finally, by concatenating the original video-level features $\boldsymbol{x_v}$, frame-level features $\boldsymbol{x_f}$ and important patch-level features $\boldsymbol{x_{ip}}$ in a coarse-to-fine manner, the multi-grained input to our pseudo-query generator is $[\boldsymbol{x_v}, \boldsymbol{x_f}, \boldsymbol{x_{ip}}] \in \mathbb{R}^{(4+m+k) \times d}$. 

\textbf{Causal attention powered query generator}.
To better reconstruct text features, we implement our generator using a causal attention transformer, initialized with the CLIP text encoder. After a linear projection, we feed $[\texttt{[bos]}, \boldsymbol{x_{v}}, \boldsymbol{x_{f}}, \boldsymbol{x_{ip}}, \texttt{[eos]}]$ into the transformer. The \texttt{[bos]} and \texttt{[eos]} tokens help facilitate the convergence during training. Finally, the output of the last transformer layer at the \texttt{[eos]} position represents the pseudo query feature. That is $\boldsymbol{t_p} = \phi_g( [\boldsymbol{x_v}, \boldsymbol{x_f}, \boldsymbol{x_{ip}}])$.


\subsection{Pseudo-Interaction Fusioner}

Then a cross-modal fusioner module is introduced to enhance the video features. We implement our fusioner with the XPool Attention module \cite{cvpr2022xpool}. It can be seen as a one-layer cross attention without the residual connection, functioning as using the text feature as guidance to perform an attention pooling over a sequence of visual features. This attention pooling mechanism can learn to highlight the relevant frames to a given textual query while suppressing frames not described. To perform a fine-grained feature fusion, we utilize pseudo-query $\boldsymbol{t_p}$
as Q, concatenated video, frame level features $[\boldsymbol{x_v}, \boldsymbol{x_f}] \in \mathbb{R}^{(4+m) \times d}$ as K\&V.
Finally, the fusioner outputs the final video representation $\boldsymbol{v} \in \mathbb{R}^d$ enhanced by ``pre-query video interaction modeling'' as follows:
\begin{equation}
\left\{ \begin{array}{ll}
\boldsymbol{v^{'}} & \leftarrow \mbox{LN}(\mbox{Attention}(Q:\boldsymbol{t_p},  ~K\&V: [\boldsymbol{x_v}, \boldsymbol{x_f}]), \\
\boldsymbol{v} & \leftarrow \mbox{LN}(\mbox{FC}(\boldsymbol{v^{'}}) + \boldsymbol{v^{'}}),
       \end{array} \right.
\end{equation}
where LN is a Layer Normalization layer \cite{arxiv2016layernorm} and FC is a fully connected network.
The learned final video representation $\boldsymbol{v}$ is then used for video retrieval in an efficient Two-Tower manner (in Sec. \ref{sec:training_inference}).
Throughout our hybrid-tower framework, the finer-grained modeling of multimodal feature extraction, generation and interaction are well concerned.




\subsection{Training and Inference}
\label{sec:training_inference}

\noindent
\textbf{Training.}
During training, we use the real text features $\boldsymbol{t}$ as supervision to constrain the reconstructed pseudo-query features $\boldsymbol{t_p}$ from videos, thereby optimizing the generator. The reconstruction loss used here is the cosine distance loss \cite{cvpr2024radio, iccv2023umt},

\begin{equation}
\mathcal{L}_{recon} = 1 - sim_{cos}(\boldsymbol{t}_p, \boldsymbol{t}),
\label{eq:recon}
\end{equation}
where $sim_{cos}()$ denotes the cosine similarity between two features. In summary, the total objective is defined by combining $\mathcal{L}_{cons}$ in Eq. \ref{eq:t2v-v2t} and $\mathcal{L}_{recon}$ in Eq. \ref{eq:recon} as:
\begin{equation}
\mathcal{L} = \mathcal{L}_{cons} + \alpha  \mathcal{L}_{recon},
\label{eq:total}
\end{equation}
where $\alpha$ is the scaling weight for the reconstruction loss. Our training process consists of two stages: (1) pretraining the generator using only reconstruction objectives while keeping the encoders and fusion module frozen; (2) fine-tuning the entire network with both contrastive and reconstruction objectives. It is worth noting that neither training stage introduces extra data. 

\noindent
\textbf{Inference.}
In inference, all video features can be precalculated and stored offline, allowing the generator to generate corresponding pseudo-queries for each video. This enables offline fine-grained feature fusion, ensuring both Two-Tower level efficiency and Single-Tower level effectiveness.

\section{Experiment}
\label{sec:exp}

\begin{table*}[ht!]
\centering

\setlength{\tabcolsep}{3pt}
\renewcommand{\arraystretch}{1.1}
\resizebox{\linewidth}{!}{
\begin{tabular}{@{}lrrrr|rrr|rrr|rrr|rrr@{}}
\toprule

\multirow{2}{*}{\textbf{Model}} & \multirow{2}{*}{\textbf{FLOPs} \small{$\downarrow$}} & \multicolumn{3}{c}{\textbf{MSRVTT-1k}} &\multicolumn{3}{c}{\textbf{MSRVTT-3k}} & \multicolumn{3}{c}{\textbf{MSVD}} & \multicolumn{3}{c}{\textbf{VATEX}} &
\multirow{2}{*}{\textbf{Mean}} \\

\cmidrule(r){3-5} \cmidrule(r){6-8} \cmidrule(r){9-11} \cmidrule(r){12-14} 
& & R@1 & R@5 & SumR & R@1 & R@5 & SumR & R@1 & R@5 & SumR & R@1 & R@5 & SumR \\
\midrule
\multicolumn{15}{@{}l}{\textit{Feature re-learning with CLIP feature}:}\\
SEA \cite{tmm2020sea} & 10.0K & 37.2 & 67.1  & 182.6 & 19.9 & 44.3  & 120.7 & 34.5 & 68.8  & 183.8 & 52.4 & 90.2  & 238.5 & 36.0 \\
W2VV++ \cite{acmmm2019w2vv++} & 2.0K & 39.4 & 68.1  & 185.6 & 23.0 & 49.0 & 132.7 & 37.8 & 71.0 &  190.4 & 55.8 & 91.2 & 243.0 & 39.0 \\
MMT \cite{eccv2020mmt} & 3.5K & 39.5 & 68.3  & 186.1 & 24.9 & 50.5 &  137.4 & 40.6 & 72.0 & 194.3 & 54.4 & 89.2  & 238.6 & 39.9 \\
LAFF \cite{eccv2022laff} & 4.0K & 45.8 & 71.5  & 199.3 & 29.1 & 54.9  & 149.8 & 45.4 & 70.6 & 200.6 & 59.1 & 91.7 & 247.1 & 44.9 \\
\midrule
\multicolumn{15}{@{}l}{\textit{\textbf{Single-Tower} CLIP-based end-to-end (visual backbone: ViT-B/32):}}\\
TS2-Net \cite{eccv2022ts2} & 6.1K & 46.7 & 72.6  & 200.5 & 29.9 & 56.4  & 153.6 & 44.6 & 75.8  & 204.9 & 61.1 & 91.5  & 248.6 & 45.6 \\
X-CLIP \cite{mm22xclip} & 220.9K & 45.3 & 73.7 & 200.8 & 31.2 & \textbf{57.4} &  \textbf{156.7} & 47.2 & 77.0  & 210.1 & 62.2 & 90.9  & 248.5 & 46.5 \\
XPool \cite{cvpr2022xpool} & 275.0K & 46.0 & 72.8 &  201.5  & -- & -- & -- & -- & -- & -- & -- & -- & -- & -- \\
DRL \cite{arxiv2022drl}  & 220.4K &  46.2 & 74.0  & 203.2  & -- & -- & -- & -- & -- & -- & -- & -- & -- & -- \\ 
ProST \cite{cvpr2023prost} & 2017.5K & 48.2 & \textbf{74.6} & \textbf{206.2} & -- & -- & -- & -- & -- & -- & 62.6 & 91.3 & 249.3 & -- \\
UCoFiA \cite{iccv2023ucofia} & 9836.7K & \textbf{49.1} & 72.1 & 202.7 & -- & -- & -- & 47.1 & 77.1 & 208.5 & 62.7 & 90.3 & 248.1 & -- \\
\midrule
\multicolumn{15}{@{}l}{\textit{\textbf{Two-Tower} CLIP-based end-to-end (visual backbone: ViT-B/32):}}\\
CLIP4Clip \cite{Neurocomputing2022clip4clip} & \textbf{0.5K} & 42.8 & 71.6  & 195.5 & 29.4 & 54.9 & 150.1 & 45.6 & 76.1  & 206.6 & 61.6 & 91.1  & 248.5 & 44.9 \\

CenterCLIP \cite{sigir2022centerclip} & 1.5K &  44.2 & 71.6 & 197.9 & -- & -- & --  & 47.3 & 76.8  & 209.7 & -- & -- & -- & -- \\

Cap4Video$^*$ \cite{cvpr2023cap4video} & 1.0K & 45.6 & 71.7  &  81.2 & -- & -- & -- & -- & -- & -- & -- & -- & -- & -- \\

PromptSwitch \cite{iccv2023promptswitch} & \textbf{0.5K} & 43.6  & 71.5  & 195.7   & -- & -- & -- & 46.3 & 75.8 & 206.6 & -- & -- & -- & -- \\
STAN \cite{cvpr2023stan} & \textbf{0.5K} & 46.9 & 72.8 & 202.5 & -- & -- & -- & -- & -- & -- & -- & -- & -- & -- \\

TeachCLIP \cite{cvpr2024teachclip} & \textbf{0.5K} & 46.8 & 74.3 & 203.7 & 30.9 & 57.1 & 156.0 & 47.4 & \textbf{77.3} &  \textbf{210.2} & 63.6 & \textbf{91.9} &  251.6 & 47.2 \\
CLIP-ViP$^*$ \cite{iclr2023clipvip} & \textbf{0.5K} & 46.0 & 72.7 & 200.6 & 30.2 & 56.0 & 155.3 & 45.1 & 74.8 & 206.5 & 62.1 & 90.2 & 249.3 & 45.9 \\
\hline
\rowcolor{mygray}
\emph{\name{}(Ours)} & \textbf{0.5K} & 48.6 & 72.8 & 203.0 & \textbf{31.8} & 57.3 & \textbf{157.1} & \textbf{47.9} & 75.9 & 207.2 & \textbf{64.0} & 91.5 & \textbf{252.1} & \textbf{48.0} \\
\bottomrule
\end{tabular}
}
\caption{\textbf{T2VR Performance of different methods on multiple datasets}. Note that we replicate existing methods using their author-provided source code where applicable, so the numbers might (slightly) differ from those in their original papers. None of the reported methods is applied by post-processing, as it is not practical in real retrieval scenarios. Cap4Video$^*$ refers to the Two-Tower version (global matching version) of Cap4Video. CLIP-ViP$^*$ refers to the version without pre-training on external large-scale datasets. Mean in the last column refers to the mean of R@1 across all datasets. FLOPs indicates the computational cost of video-text matching per pair during inference.}
\vspace{-3pt}
\label{tab:alldatasets}
\end{table*}

\subsection{Experimental Setup}

\textbf{Datasets.} 
We conduct experiments on four public datasets for evaluation, including: (1) \textbf{MSRVTT} \cite{cvpr2016msrvtt}, which consists of 10K YouTube video clips, each paired with 20 human-labeled captions. Following \cite{cvpr2024teachclip}, we use two splits: MSRVTT-1k \cite{eccv2018joint}, which contains 9K videos for training and 1K video-text pairs for testing; and MSRVTT-3k \cite{cvpr2016msrvtt}, the original split, which consists of nearly 7K videos for training and 3K video-text pairs for testing. The MSRVTT-1k split is used as the primary dataset for our ablation study. (2) \textbf{MSVD} \cite{acl2011msvd} contains 1,970 videos, with 80K captions, averaging 40 captions per video. We follow the official split of 1,200 and 670 as the train and test set, respectively. (3) \textbf{VATEX} \cite{iccv2019vatex} is comprised of around 35K videos, each with multiple annotations. Due to some unavailable videos, we follow the split in \cite{cvpr2024teachclip}, which includes 23,896 videos for training, 1,375 for validation, and 1,398 for testing. (4) \textbf{DiDeMo} \cite{emnlp2018didemo} consists of nearly 10k video clips and 40k captions in total. We adopt the official data split which has 8,394 video clips for training, 1,065 for validation, and 1,004 for testing.

\textbf{Evaluation criteria.}
Following previous works, we utilize the most common rank-based metrics, \ie Recall at top-k (R@k, k=1, 5, 10), SumR
(R1+R5+R10) and mean rank (MnR) to evaluate models.


\textbf{Implementation Details.} Experiments are conducted on 8 NVIDIA 3090 GPUs. The default setting is as follows, unless otherwise stated. We use the CLIP model initialized by OpenAI-released version. The learning rate is set to 9e-5 during the pseudo-query generation pertaining stage and 1e-6 during full fine-tuning. The maximum length of frame / word tokens is set to 12/50, respectively, with a batch size of 128. For DiDeMo’s longer clips, we follow \cite{iclr2023clipvip} and use 32/64 tokens. The $\alpha$ in Eq. \ref{eq:total} is set to 2 and the top-k in informativeness token selection is set to top-16 for optimal performance.

\subsection{Performance Comparison}

\textbf{Baselines.}
Both feature re-learning based methods and CLIP-based end-to-end methods are compared. For the purpose of fair comparison and reproducibility of research, we only include feature re-learning methods with CLIP feature and those open-sourced methods as follows:
\begin{itemize}
    \item \textit{Feature re-learning}: SEA \cite{tmm2020sea}, W2VV++ \cite{acmmm2019w2vv++}, MMT \cite{eccv2020mmt} and LAFF \cite{eccv2022laff}.
    \item \textit{Single-Tower CLIP-based end-to-end}: TS2-Net \cite{eccv2022ts2}, X-CLIP \cite{mm22xclip}, XPool \cite{cvpr2022xpool}, DRL \cite{arxiv2022drl}, ProST \cite{cvpr2023prost} and UCoFiA \cite{iccv2023ucofia}.
    \item \textit{Two-Tower CLIP-based end-to-end}: CLIP4clip \cite{Neurocomputing2022clip4clip}, CenterCLIP \cite{sigir2022centerclip}, Cap4Video \cite{cvpr2023cap4video}, PromptSwitch \cite{iccv2023promptswitch},
    STAN \cite{cvpr2023stan}, TeachCLIP \cite{cvpr2024teachclip} and our backbone baseline CLIP-ViP \cite{iclr2023clipvip}.
\end{itemize}
Among the end-to-end methods, only some of them have results across all datasets, see \tabref{tab:alldatasets}.

\textbf{Efficiency comparison.} Following \cite{cvpr2024teachclip}, we calculate the FLOPs\footnote{\href{https://github.com/sovrasov/flops-counter.pytorch}{https://github.com/sovrasov/flops-counter.pytorch}} required for per video-text matching during serving, \ie video features can be pre-extracted offline and stored in advance. As \tabref{tab:alldatasets} shows, our method achieves the best efficiency of 0.5K FLOPs per matching by keeping the retrieval process in a Two-Tower manner. We also report additional results from both online and offline perspectives in \tabref{tab:overhead}. The goal of our work is to improve online-stage performance, and given the strong gains achieved, we consider the offline overhead acceptable.

\begin{table}[tbp!]
\centering
\setlength{\tabcolsep}{3pt} 
\renewcommand{\arraystretch}{1.1} 
\resizebox{\linewidth}{!}{
\begin{tabular}{@{}clrrrr@{}}
\toprule
\textbf{\#} & \textbf{Setup} & \textbf{R@1} \small{$\uparrow$} & \textbf{R@5} \small{$\uparrow$} & \textbf{R@10} \small{$\uparrow$} & \textbf{MnR} \small{$\downarrow$} \\
\midrule
\rowcolor{mygray}
0 & Full-setup & \textbf{48.6} & \textbf{72.8} & 81.6 & 15.4 \\
\multicolumn{4}{@{}l}{\quad \textit{Loss Functions:}} \\
1 & $w/o$ $\mathcal{L}_{recon}$ & 45.8 & 71.8 & 81.5 & 16.7   \\
[3pt]
\multicolumn{4}{@{}l}{\quad \textit{Input of Generator:}} \\
2 & $\phi_g(\boldsymbol{x_v})$ as $\boldsymbol{t_p}$ & 46.3 & 72.0 & 81.4 & 16.1 \\
3 & $\phi_g([\boldsymbol{x_v}, \boldsymbol{x_f}])$ as $\boldsymbol{t_p}$ & 47.5 & 71.8 & 81.5 & 15.9 \\
4 & $\phi_g([\boldsymbol{x_f}, \boldsymbol{x_{ip}}])$ as $\boldsymbol{t_p}$ & 48.1 & 71.8 & 81.4 & 15.6 \\
[3pt]
\multicolumn{4}{@{}l}{\quad \textit{Architecture of Generator:}} \\
5 & Causal Attn. $\rightarrow$ MLP & 44.1 & 70.5 & 80.2 & 17.1   \\
6 & Causal Attn. $\rightarrow$ Q-former \cite{icml2023blip2, eccv2020detr} & 47.6 & 72.5 & 81.0 & 15.5  \\
[3pt]
\multicolumn{4}{@{}l}{\quad \textit{Architecture of Fusioner:}} \\
7 & XPool $\rightarrow$ Cross Attn. & 42.2 & 70.0 & 80.6 & 16.2 \\
8 & XPool $\rightarrow$ Co Attn.\cite{nips2019vilbert} & 44.5 & 71.6 & 80.3 & 17.0 \\
[3pt]
\multicolumn{4}{@{}l}{\quad \textit{Form of Pseudo-Query:}} \\
9 & Captions (ShareGPT4Video \cite{nips2025sharegpt4video}) & 43.8 & 69.2 & 78.5 & 18.5 \\
10 & Captions (m-PLUG2 \cite{icml2023mplug2}) & 47.4 & 71.7 & \textbf{82.0} & \textbf{15.3} \\

\bottomrule
\end{tabular}
}
\caption{\textbf{Ablation Study of \name.} Backbone: CLIP-ViT-B/32. Dataset: MSRVTT-1k.}
\vspace{-5pt}
\label{tab:abl-arch}
\end{table}

\textbf{Effectiveness comparison.}
\tabref{tab:alldatasets} presents the performance of various methods on multiple video retrieval benchmarks. As observed, feature re-learning models (top section) are notably inferior to CLIP-based end-to-end methods. 
As for the end-to-end methods, our method \name{} stands out with a mean R@1 of 48.0, clearly outperforming its Two-Tower counterparts.
In particular, the backbone model of \name{}, CLIP-ViP, is enhanced, yielding R@1 improvements of 2.6 on MSRVTT-1k, 1.6 on MSRVTT-3k, 2.8 on MSVD, and 1.9 on VATEX, thereby surpassing the previous state-of-the-art TeachCLIP.
Moreover, although Single-Tower models such as ProST and UCoFiA achieve slight gains in R@1 (0.5) and R@5 (1.8), they incur 4,035-fold and 19,673-fold higher inference costs, respectively, compared to \name{}.
Thus, the exceptionally low computational cost of \name{} (0.5K FLOPs) ensures remarkable efficiency, making it highly suitable for real-time retrieval applications. These observations, along with its robust performance across diverse datasets, demonstrate that \name{} balances high retrieval effectiveness with outstanding computational efficiency.


We also evaluate our model with different backbone scales and architectures. As shown in \tabref{tab:backbones}, whether we vary size (Setup-1/2 \emph{vs} Setup-4/5) or architecture (Setup-0/1 \emph{vs} Setup-3/4), our model consistently outperforms the baselines. We find that performance on DiDeMo is poor when CLIP-ViP is used as the backbone. This counterintuitive result stems from DiDeMo’s long videos (32 frames), which force CLIP-ViP’s proxy attention to handle 32 × 49 tokens simultaneously, which is very challenging. After switching the backbone to CLIP4Clip, the absolute R@1 on DiDeMo increases from 41.2 to 44.1.

\begin{table}[tp!]
    \centering
    \resizebox{0.9\linewidth}{!}{
        \begin{tabular}{@{}lrrrrr@{}}
        \toprule
        \multirow{3}{*}{\textbf{Model}}  & 
        \multicolumn{2}{c}{\textbf{Offline Stage}} & 
        \multicolumn{2}{c}{\textbf{Online Stage}} & \multirow{3}{*}{\makecell{\textbf{R@1} \\ \textbf{($\uparrow$)}}} \\
        \cmidrule(lr){2-3} \cmidrule(lr){4-5}
        ~ & \makecell{Video feature\\extraction ($\downarrow$)\\(FLOPS)} & \makecell{Model ($\downarrow$)\\parameters\\(M)} 
          & \makecell{Per video-text\\matching ($\downarrow$)\\(FLOPS)} & \makecell{Video feature\\storage ($\downarrow$)\\(KB)} & \\
        \midrule
        ProST & 53.89G & 180.4 & 2017.5K & 294 & 48.2 \\
        TeachCLIP & \textbf{53.65G} & \textbf{164.2} & \textbf{0.5K} & \textbf{2} & 46.8 \\
        \rowcolor{mygray}
        \emph{F-Pig(Ours)} & 54.90G & 191.0 & \textbf{0.5K} & \textbf{2} & \textbf{48.6} \\
        \bottomrule
        \end{tabular}
    }
    \caption{\textbf{Online/offline overhead.} Dataset: MSRVTT-1k.}
    \label{tab:overhead}
\end{table}

\begin{table}[tbp!]
    \centering
    \small
    \setlength{\tabcolsep}{3pt}
    \resizebox{\linewidth}{!}{%
    \begin{tabular}{@{}clrrr rrr@{}}
    \toprule
    \multirow{2}{*}{\textbf{\#}} & \multirow{2}{*}{\textbf{Setup}} 
      & \multicolumn{3}{c}{\textbf{MSRVTT-1k}} 
      & \multicolumn{3}{c}{\textbf{DiDeMo}} \\
    \cmidrule(lr){3-5} \cmidrule(lr){6-8}
    ~ & ~ & R@1\small{$\uparrow$} & R@5\small{$\uparrow$} & R@10\small{$\uparrow$} 
        & R@1\small{$\uparrow$} & R@5\small{$\uparrow$} & R@10\small{$\uparrow$} \\
    \midrule 
    \multicolumn{8}{@{}l}{\quad \textit{Baselines:}} \\
    0 & CLIP4Clip (B/32)         & 42.8 & 71.6 & 81.1 & 42.0 & 69.0 & 78.2 \\
    1 & CLIP-ViP (B/32)          & 46.0 & 72.7 & 81.8 & 39.5 & 70.1 & 76.9 \\
    2 & CLIP-ViP (B/16)          & 49.4 & 73.6 & 84.0 & 41.8 & 71.0 & 80.3 \\
    \midrule
    \multicolumn{8}{@{}l}{\quad \textit{Backbone of \name{}:}} \\
    3 & \name{} $w$ CLIP4Clip (B/32) & 45.1 & 72.1 & 81.5 & \textbf{44.1} & \textbf{71.9} & 81.3 \\
    4 & \name{} $w$ CLIP-ViP (B/32)  & 48.6 & 72.8 & 81.6 & 41.2 & 67.9 & 80.5 \\
    5 & \name{} $w$ CLIP-ViP (B/16)  & \textbf{51.2} & \textbf{75.1} & \textbf{84.5} & 43.3 & 70.8 & \textbf{81.9} \\
    \bottomrule
    \end{tabular}%
    }
    \caption{\textbf{Performance of \name{} with different backbones and sizes.} Here, B/32 and B/16 refer to CLIP-ViT-B/32 and CLIP-ViT-B/16 backbones, respectively.}
    \vspace{-5pt}
    \label{tab:backbones}
\end{table}

\subsection{Ablation Studies}
The essence of our \name{} is the pseudo-query generator and the pseudo-interaction fusioner. Hence, \name{} needs to be evaluated along multiple dimensions, including choice of generator's input, architecture of generator, the architecture of fusioner, the form of pseudo-query.


\textbf{Choice of Input Visual Tokens.} The input visual features for the generator include video-level ($\boldsymbol{x_v}$), frame-level ($\boldsymbol{x_f}$), and informative patch-level ($\boldsymbol{x_{ip}}$) features. 

As shown in \tabref{tab:abl-arch}, \name{} consistently surpasses the baseline regardless of feature granularity, achieving R@1 improvements from 0.3 to 2.6. Notably, informative patch-level features yield the greatest gain, while the best results are obtained when using all multi-grained inputs. Without top-k selection, feeding all patch-level features causes out-of-memory issues; for example, a 12-frame video with CLIP-ViT-B/32 yields $12 \times 49 = 588$ patch features. This highlights the necessity of reducing patch-level feature numbers.

\textbf{Architecture of generator.}
As \tabref{tab:abl-arch} shows, switching from the causal attention transformer initialized by CLIP’s text encoder to other architectures (Setup-5, Setup-6) degrades performance.
Due to its sequence modeling ability, the causal attention transformer naturally handles our coarse-to-fine multi-grained visual inputs.




\textbf{Architecture of fusioner.} The effectiveness of different fusioner architectures is evaluated in \tabref{tab:abl-arch}. XPool Attention (Setup-0) outperforms Cross Attention (Setup-7) and Co Attention (Setup-8) across all metrics. The latter two lag because of their residual connection, which directly adds the weak pseudo-query feature to the final video representation, thereby weakening its quality. By contrast, our design captures fine-grained text–frame dependencies more effectively.

\textbf{Form of pseudo-query.}
How to generate effective pseudo-queries? There are two straightforward solutions: (1) utilizing an external video captioner to generate raw auxiliary captions as pseudo-queries \cite{cvpr2023cap4video}; (2) training a pseudo-query generator aligned with the text features of real queries (our method). We employ two powerful video captioning models, ShareGPT4Video \cite{nips2025sharegpt4video} and m-plug2 \cite{icml2023mplug2}, to generate auxiliary captions on MSRVTT-1ka. The fine-tuned CLIP text encoder is then used to extract auxiliary text features as pseudo-query features. As shown in \tabref{tab:abl-arch}, our solution (Setup-0) outperforms the two caption-based solutions (Setup-9, Setup-10), demonstrating that our trained pseudo-query generator produces better text features.

\textbf{Reconstruction Loss.} Removing text feature supervision significantly harms our model’s performance (Setup-1). Without the guidance of our reconstruction loss, the pseudo-queries degenerate into meaningless noise.

\begin{table}[tbp!]
\centering
\setlength{\tabcolsep}{2pt}
\resizebox{0.7\linewidth}{!}{
\begin{tabular}{@{}lrrrr@{}}
\toprule
 \textbf{Model} & \textbf{MSVD} & \textbf{VATEX} & \textbf{DiDeMo} & \textbf{Mean} \\ \hline 
CLIP4Clip & 43.7 & 48.5
 & 30.3 & 40.8 \\
 X-CLIP & 44.8 & 48.9
 & \textbf{33.5} & \textbf{42.4} \\
TeachCLIP & 44.0 & 49.1 & 31.3
 & 41.5 \\
 \rowcolor{mygray}
F-Pig & \textbf{44.9} & \textbf{49.8} & 30.9 & 42.0 \\
\bottomrule
\end{tabular}
}

\caption{\textbf{Cross-dataset results}. Backbone: CLIP-ViT-B/32. Training data: MSRVTT-1k.}
\label{tab:cross-dataset}
\vspace{-10pt}
\end{table}

\textbf{Cross dataset evaluation.}
To check whether the generated pseudo-queries are restricted to a specific dataset domain, we evaluate our method in out-of-domain retrieval settings \cite{cvpr2024teachclip, iccv2023diffusionret}. As shown in \tabref{tab:cross-dataset}, we directly test our model across different datasets. \name{} achieves the best R@1 performance on MSVD and VATEX, demonstrating the strong generalization ability of our approach.

\subsection{Qualitative Analysis.}
\begin{figure}[tp!]
    \centering
    \includegraphics[width=0.8\linewidth]{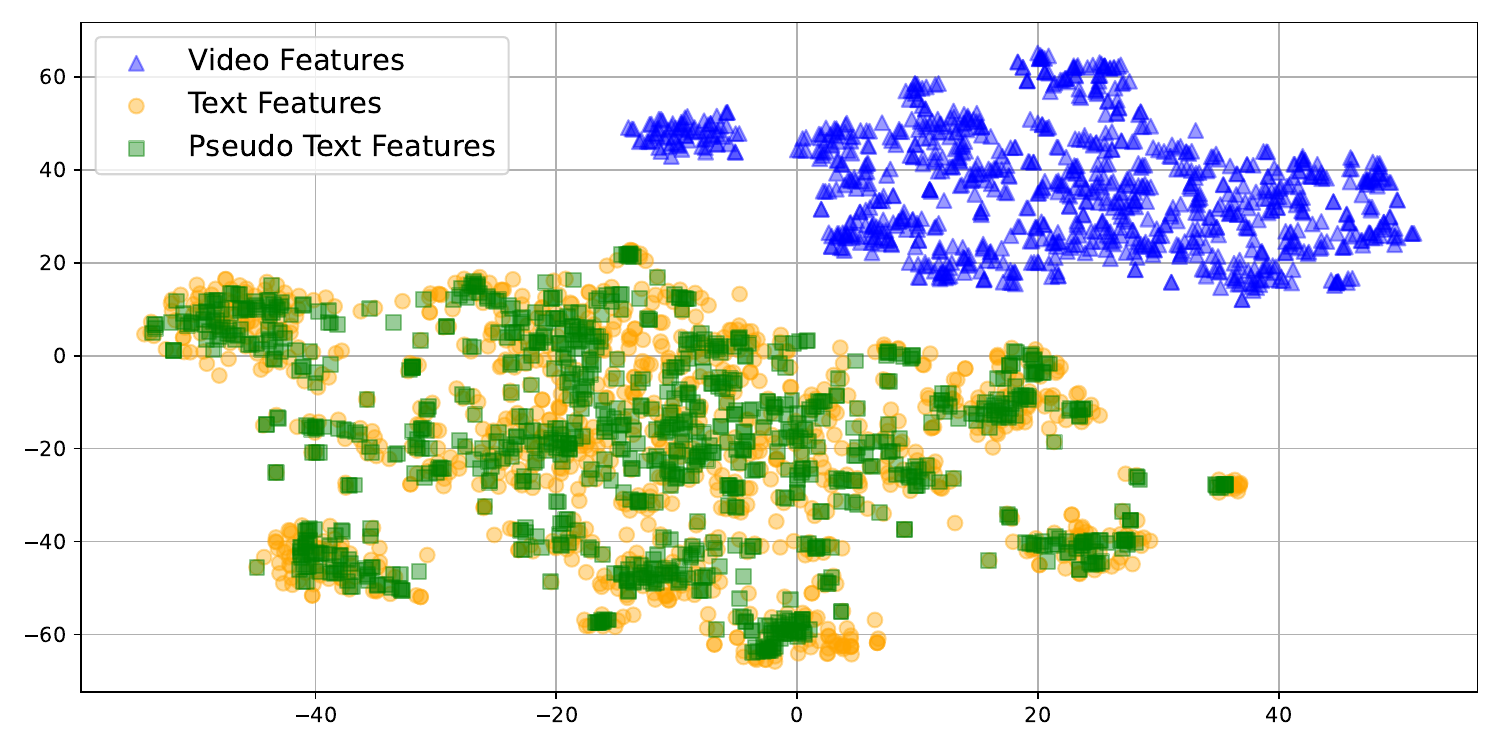}
    \caption{t-SNE visualization of the embedding space generated by our \name{} on the MSRVTT-1k test set.}
    \label{fig:tsne}
\end{figure}

\textbf{Embedding space visualization.} \figref{fig:tsne} shows the t-SNE plot of the joint embedding space. The pseudo-text features generated solely from video blend seamlessly with the real text features, demonstrating that our method learns meaningful pseudo-query embeddings aligned with the distribution of real text queries. Consequently, these pseudo-queries effectively guide the video features toward a finer-grained visual representation.

\begin{figure}[tb!]
  \centering

  \begin{subfigure}[t]{0.49\linewidth}
    \centering
    \includegraphics[width=\textwidth]{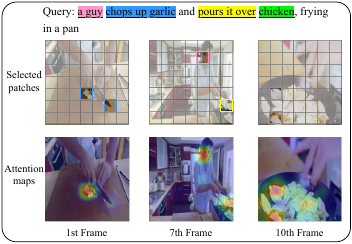}
    \label{fig:example0}
  \end{subfigure}\hfill
  \begin{subfigure}[t]{0.49\linewidth}
    \centering
    \includegraphics[width=\textwidth]{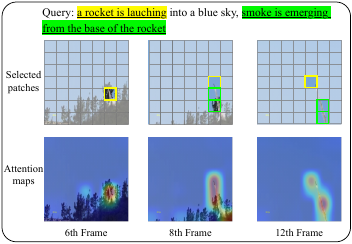}
    \label{fig:example2}
  \end{subfigure}
  \\[-4pt]   
  \begin{subfigure}[t]{0.49\linewidth}
    \centering
    \includegraphics[width=\textwidth]{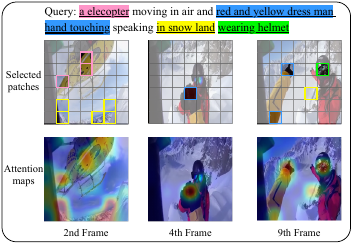}
    \label{fig:example1}
  \end{subfigure}\hfill
  \begin{subfigure}[t]{0.49\linewidth}
    \centering
    \includegraphics[width=\textwidth]{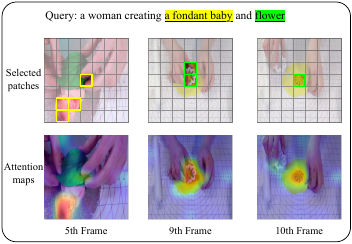}
    \label{fig:example3}
  \end{subfigure}

  \vspace{-8pt}
  \caption{\textbf{Qualitative results of our ITS module.} Top row: patch tokens selected by our ITS module, which are then fed into the pseudo-query generator. Bottom row: frame attention maps. The corresponding patch tokens
   with the phrases in the query are highlighted in the same color.}
   \vspace{-8pt}
  \label{fig:examples}
\end{figure}

\textbf{Qualitative Examples.} \figref{fig:examples} presents qualitative results on MSRVTT-1k. Leveraging our ITS module, we retain the patches that contain the fine-grained information needed to generate discriminative pseudo-queries.

\section{Conclusion}


We propose a Hybrid-Tower framework for Text-to-Video Retrieval (T2VR) that balances the effectiveness of Single-Tower models with the efficiency of Two-Tower models. Our method, Fine-grained Pseudo-query Interaction and Generation (\name{}), introduces a pseudo-query generator and fusioner to enable fine-grained feature fusion before the arrival of real queries. This achieves near-SOTA accuracy while maintaining efficiency. Extensive experiments validate its superiority, highlighting its potential to inform future research in T2VR.

\paragraph{Acknowledgments}
This work was supported by NSFC (62172420), Tencent Marketing Solution Rhino-Bird Focused Research Program and the Young Elite Scientists Sponsorship Program by CAST (2023QNRC001).

{
    \small
    \bibliographystyle{ieeenat_fullname}
    \bibliography{main}
}

\end{document}